%% file: main.tex
\documentclass[sigconf]{acmart}

\AtBeginDocument{%
  }

\setcopyright{acmcopyright}
\copyrightyear{2023}
\acmYear{2023}
\acmDOI{XXXXXXX.XXXXXXX}
\newcommand{\hide}[1]{}

\usepackage{caption}
\usepackage{subfigure}
\usepackage{multirow}
\usepackage{diagbox}
\usepackage{url}
\usepackage[ruled,linesnumbered]{algorithm2e}
\usepackage{algorithmic}

\begin{document}


%\title{Graph Regularizer Augmentation for Data Recovery from Multiple Aggregated Views}

\title{Towards Aligned Canonical Correlation Analysis: Preliminary Formulation and Proof-of-Concept Results}


\author{Biqian Cheng}
\affiliation{%
  \institution{UC Riverside}
 \city{Riverside, CA}
  \country{USA}
  }
\email{bchen158@ucr.edu}

\author{Evangelos E. Papalexakis}
\affiliation{%
 \institution{UC Riverside}
 \city{Riverside, CA}
 \country{USA}
}
 \email{epapalex@cs.ucr.edu}
 
 \author{Jia Chen}
\affiliation{%
  \institution{UC Riverside}
  \city{Riverside, CA}
  \country{USA}
}
\email{jiac@ucr.edu}


\begin{abstract}
Canonical Correlation Analysis (CCA) has been a tried and tested analytical model which seeks to jointly embed two or more views of the data in a maximally correlated latent space. In addition to a powerful data mining tool, CCA has many implications for cutting-edge self-supervised representation learning approaches, as one can cast a number of recent approaches as variants of that model. In this work we explore what happens when a fundamental assumption of that model breaks: what if the alignment across different views of the data is unknown? Typically, we would first attempt to align the different views, and subsequently apply CCA or any other model in order to embed the data views. Can we do better if we align and embed at the same time? In this work, we seek to jointly solve the alignment and embedding by formulating and solving both problems under the same umbrella of Aligned Canonical Correlation Analysis (ACCA). We present a preliminary formulation and alternating optimization algorithm and proof-of-concept results.

\end{abstract}


\maketitle

\section{Introduction}
\label{sec-intro}
\input{Sections/introduction}

\section{Background}
\label{sec-backgound}
\input{Sections/background}

%\section{Tensors and decompositions}
%\label{sec-tensorsdecomp}
%\input{Sections/TensorsandDecompositions}

\section{Proposed Method}
\label{sec-methods}
\input{Sections/methods}

\section{Experimental Evaluation}
\label{sec-Experimentalevaluations}
\input{Sections/Experimentalevaluations}

\section{Conclusion \& Future Work}
\label{sec-Conclusion}
\input{Sections/conclusion}

\begin{acks}
The authors would like to thank Yunshu Wu for initial discussions.
Research was supported by the National Science under CAREER grant no. IIS 2046086 and CREST Center for Multidisciplinary Research Excellence in Cyber-Physical Infrastructure Systems (MECIS) grant no. 2112650 and by the US Department of Transportation under University Transportation Center (UTC) on Railway Safety. Any opinions, findings, and conclusions or recommendations expressed in this material are those of the author(s) and do not necessarily reflect the views of the funding parties.

\end{acks}

\bibliographystyle{plainnat}
%{\footnotesize
\bibliography{main.bib}
%}
\end{document}

%% file: Sections/introduction.tex
%1- Problem motivation 
%why is interesting 
%\reminder{Para 1: Motivate CCA with examples of two bipartite graphs (user, product) and (user, video) or generally two different feature spaces (text vs. image or whatever). Say that CCA is a powerful way of finding a common latent space for those heterogeneous data \cite{}. Then say that even though CCA has been around for a long time, it is still extremely relevant as cutting-edge self-supervised representation learning techniques such as Barlow Twins \cite{zbontar2021barlow,bielak2022graph,zhang2021canonical,balestriero2023cookbook} can be essentially seen as variations of CCA.}
Canonical Correlation Analysis (CCA) \cite{harold1936relations,kettenring1971canonical} is a classical model which, given two different views of the same set of entities, e.g., two different bipartite graphs of (user, product) and (user, video) interactions or different feature representations for those entities in general, seeks to project those entities (users) in a low-dimensional space where the different projected views are maximally correlated. Essentially, CCA can jointly embed heterogeneous datasets in a common low-dimensional space, as it can be extended to more than two views \cite{chen2019graph,chen2022unsupervised}. Even though CCA has been in and out of the spotlight for many decades and has been around for quite a long time, it is still extremely relevant, not only as a standalone data mining tool, but also due to the fact that cutting-edge self-supervised representation learning techniques, such as Barlow Twins \cite{zbontar2021barlow,bielak2022graph,zhang2021canonical,balestriero2023cookbook}, can be essentially seen as variations of CCA, where the goal is to embed two views of the data (the original view and the augmentation) in a latent space where correlation is maximized.

Traditionally, in CCA-style analysis, we assume that entities across views have one-to-one correspondence across the two or more views of the data, and there is a wealth of algorithms that study different formulations for solving the problem of projecting those views in that desired maximally correlated space, both linearly and non-linearly \cite{andrew2013deep}. What if, however, this one-to-one correspondence is unknown? In this case, we are faced with two problems: (1) entity alignment and (2) CCA embedding. Motivated by recent results \cite{wu2022tenalign} in the related problem of misaligned joint tensor factorization, it turns out that formulating and solving the alignment and embedding problems jointly yields better results than solving each problem separately in multiple steps, as it appears that the two sub-problems work synergistically to produce better quality alignment and embeddings. In this work, we explore this type of formulation for CCA and we propose a new formulation, the Aligned Canonical Correlation Analysis (ACCA), where we seek to jointly compute the alignment and the embedding space.

%\reminder{Para 2: CCA assumes that there is a known alignment of the data points (users). What if this alignment is unknown? In this paper we propose a new formulation of the model, the Aligned Canonical Correlation Analysis (ACCA), where we jointly compute the alignment and the embedding space. We have recently shown that jointly formulating this problem for aligned tensor factorization (CITE TenAlign)\cite{wu2022tenalign} makes the alignment and factorization problems to work synergistically, producing better quality alignment and embeddings than doing so in multiple steps. In this paper we endavor to explore this for CCA.}

The closest formulation to our proposed model is found in \cite{sahbi2018learning} where the author is considering linear transformation of the two views in CCA, however, is not seeking to recover the precise alignment matrix as our formulation does. In our on-going work we will consider scenarios where we can fairly compare the two formulations and understand pros and cons for either one.

The list of contributions in this preliminary work are:
\begin{itemize}
    \item {\bf Novel Formulation}: We propose the Aligned Canonical Correlation Analysis (ACCA) model, which seeks to jointly identify the best entity alingment and latent embedding for the dataset views.
    \item{\bf Proof of Concept}: We derive an Alternating Optimization algorithm and show preliminary results for solving the problem, demonstrating the feasibility of our effort.
\end{itemize}

%% file: Sections/background.tex
%\reminder{Describe the basic version of CCA we are looking into}

Canonical correlation analysis (CCA) is a powerful tool to learn the shared latent components of two datasets by projecting them to the same space and enforcing the similarity of the projected data. Given two centered and aligned datasets $\mathbf{X}\in \mathbb{R}^{D_x \times N}$ and $\mathbf{Y}\in \mathbb{R}^{Dy \times N}$ where $N$ is the number of samples, $D_x$ and $D_y$ represent the dimensions of $\mathbf{X}$ and $\mathbf{Y}$, respectively, one popular CCA formulation is seeking for the two projection matrices $\mathbf{U}\in\mathbb{R}^{d\times D_x}$ and $\mathbf{V}\in\mathbb{R}^{d\times D_y}$ with $d\ll \min (D_x,\,D_y)$, and shared representation/embedding $\mathbf{S}\in\mathbb{R}^{d \times N}$ by solving the following  problem
% \begin{align}
% 	\min_{\mathbf{U},\mathbf{V}, \mathbf{S}}  \quad&  \left \|\mathbf{U}{\mathbf{X}} -\mathbf{S}\right\|_F^2 + \left\|\mathbf{V}{\mathbf{Y}} -\mathbf{S}\right\|_F^2\label{eq:cca}
%  \end{align}
\begin{equation}
    {\textbf{min}}_{\mathbf{U,V,S}} || \mathbf{UX} - \mathbf{S} ||^{2}_{F} + || \mathbf{VY} - \mathbf{S} ||^{2}_{F}
 \label{eq:cca}
\end{equation}
\noindent under the constraint that $\mathbf{S}\mathbf{S}^\top=\mathbf{I}$ which avoids the trivial solution, i.e., $\mathbf{U}=\mathbf{0}$, $\mathbf{V}=\mathbf{0}$, and $\mathbf{S}=\mathbf{0}$, and ensures the $d$ latent components assembled in the rows of $\mathbf{S}$ are uncorrelated to each other. Here, the symbols $\top$ and $\|\cdot\|_F$ respectively stand for matrix transpose and Frobenius norm operators, and $\mathbf{I}$ is identity matrix with the suitable size. The minimization problem in Eq. \eqref{eq:cca} admits global optimal solution: the rows of $\mathbf{S}$ are the $d$ eigenvectors corresponding to the top-$d$ eigenvalues of $\mathbf{X}^\top(\mathbf{X}\mathbf{X}^\top)^{-1}\mathbf{X}+\mathbf{Y}^\top(\mathbf{Y}\mathbf{Y}^\top)^{-1}\mathbf{Y}$ with $(\cdot)^{-1}$ denoting the matrix inverse operator, $\mathbf{U} = \mathbf{S}\mathbf{X}^\top
(\mathbf{X}\mathbf{X}^\top)^{-1}$, and $\mathbf{V} =\mathbf{S}\mathbf{Y}^\top (\mathbf{Y}\mathbf{Y}^\top)^{-1}$, e.g., \cite{harold1936relations}.

%% file: Sections/methods.tex
%\reminder{First describe the formulation, then show how we solve the optimization problem}

The traditional CCA formulations require the entities/samples from both $\mathbf{X}$ and $\mathbf{Y}$ to be aligned, i.e., the $i$-th columns of  $\mathbf{X}$ and $\mathbf{Y}$ correspond to the two views/observations of the same latent data sample which is the groundtruth of the $i$-th
column of $\mathbf{S}$. However, if such entity alignment is imperfect, CCA is not able to learn the meaningful latent representations shared by two datasets. Toward this end, we propose a novel model, namely \emph{aligned canonical correlation analysis} (
ACCA), to jointly learn the latent representations of two views and recover the entity alignment between the two views.

\subsection{Proposed Formulation for ACCA}
Consider two centered datasets $\mathbf{X}\in \mathbb{R}^{D_x \times N}$ and $\mathbf{Y}\in \mathbb{R}^{Dy \times N}$ are two views in one dataset, and the alignment between the columns of the two views, denoted as $\bar{\mathbf{P}}\in\mathbf{R}^{N\times N}$, is unknown. Our goal is to learn the latent component representation $\mathbf{S}$ and predict the alignment matrix $\bar{\mathbf{P}}$ iteratively. Let's denote the estimation of $\bar{\mathbf{P}}$ as ${\mathbf{P}}\in\mathbf{R}^{N\times N}$. Theoretically, $\mathbf{P}$ should be a (binary) permutation matrix, and the sum of row or column is one, which shows that $\mathbf{P}$ is an orthogonal matrix. Mathematically, we will minimize $\left \|\mathbf{U}{\mathbf{X}} -\mathbf{S}\right\|_F^2 + \left\|\mathbf{V}{\mathbf{Y}}\mathbf{P} -\mathbf{S}\right\|_F^2$ under the constraints of $\mathbf{P}$ as well as the constraints from CCA, i.e., $\mathbf{SS}^\top =\mathbf{I}$. To address the computational limitation in such optimization problem, we define a list of constraints to describe different aspects of a permutation matrix instead of enforcing it to be one, for a tractable optimization solution. By relaxing the constraints on $\mathbf{P}$, the optimization formulation of our proposed ACCA is shown as:
\begin{align}
	\min_{\mathbf{U},\mathbf{V}, \mathbf{S},\mathbf{P}}   & \left \|\mathbf{U}{\mathbf{X}} -\mathbf{S}\right\|_F^2 + \left\|\mathbf{V}{\mathbf{Y}}\mathbf{P} -\mathbf{S}\right\|_F^2+\gamma_1\|\mathbf{P}\mathbf{P}^\top -\mathbf{I}\|_F^2+\gamma_2\|\mathbf{P}^\top\mathbf{P} -\mathbf{I}\|_F^2\label{eq:acca_loss}\\
  \text {S. T. } \quad&\mathbf{SS}^\top =\mathbf{I}, \text{(uncorrelatedness)}\\ 
 &0\le p_{i,j}\le 1, \forall i, j, \text{(nonnegativity)}\label{eq:con1}\\
 &\sum_{j=1}^N p_{i,j}=1, \forall i, \text{(row-wise sum )}\label{eq:con2}\\
 &H(\mathbf{p}_i)\le \lambda, \forall i \text{( entropy)}\label{eq:acca}
	\end{align}
 where $p_{i,j}$ is the $(i,j)$-th entry of $\mathbf{P}$, $\mathbf{p}_i$ is the $i$-th row of $\mathbf{P}$, $H(\mathbf{p}_i)$ is the entropy of $\mathbf{p}_i$ by viewing the $N$ entries of $\mathbf{p}_i$ as a discrete probability distribution, and the hyperparamters $\gamma_1$, $\gamma_2$, and $\lambda$ are nonnegative. Enforcing the low entropy of $\mathbf{p}_i$ guarantees that the distribution is closed to a deterministic distribution, and the second and third terms in Eq. \eqref{eq:acca_loss} will promote the orthogonality of $\mathbf{P}$

\subsection{Alternating Optimization for ACCA}
To solve the ACCA formulation, as the solution of CCA is dependent on the estimation of permutation matrix, and vice versa, we  adopt alternating optimization method, shown in Algorithm \ref{alg:our}. 

%\reminder{Biqian, what's the corresponding solver of fmincon in Python, add citation, modify this accordingly in Algorithm 1}

\begin{algorithm}[t]
	\caption{Aligned Canonical Correlation Analysis}
	\label{alg:our}
	\begin{algorithmic}[1]
		\STATE {\bfseries Input:}
		centered datasets $\mathbf{X}$ and $\mathbf{Y}$; dimension of the latent representation $d$; hyperparameters $\gamma_1$, $\gamma_2$, and $\lambda$;  and initialization of $\mathbf{P}$.
		\STATE {\bfseries Repeat}\label{step:4} \newline  
		Update $\mathbf{S}$: the rows of $\mathbf{S}$ are the $d$ eigenvectors corresponding to the top-$d$ eigenvalues of $\mathbf{X}^\top(\mathbf{X}\mathbf{X}^\top)^{-1}\mathbf{X}+(\mathbf{YP})^\top(\mathbf{YP}\mathbf{P}^\top\mathbf{Y}^\top)^{-1}\mathbf{YP}$.
		 \newline  
		Update $\mathbf{U}$: $\mathbf{U} = \mathbf{S}\mathbf{X}^\top
(\mathbf{X}\mathbf{X}^\top)^{-1}$.
		 \newline  
		Update $\mathbf{V}$: $\mathbf{V} =\mathbf{S}(\mathbf{YP})^\top(\mathbf{YP}\mathbf{P}^\top\mathbf{Y}^\top)^{-1}$.
  \newline  
		Update $\mathbf{P}$ using \emph{scipy.optimize.minimize} solver.
		\STATE {\bfseries Until} the objective Eq. \eqref{eq:acca_loss} is below a threshold or the number of iterations is beyond another threshold. 
		\STATE {\bfseries Output:}  $\mathbf{U, V, S, P}$.
	\end{algorithmic}
\end{algorithm}

%% file: Sections/Experimentalevaluations.tex
To validate the effectiveness of our proposed model ACCA, we will generate synthetic data with groundtruth $\mathbf{P}$ and investigate the performance of estimated $\mathbf{P}$ in terms of the matching accuracy between the entities in $\mathbf{X}$ and $\mathbf{Y}$. In all numerical tests, we set the hyperparameters $\gamma_1$ and $\gamma_2$ to be $0.0001$. The initial $\mathbf{P}$ is obtained by solving the optimal matching directly using $\mathbf{X}$ and $\mathbf{Y}$ without considering the canonical correlation between the two datasets, i.e., solving the following minimization problem 
\begin{align}
\min_{\mathbf{P}}   & \left \|{\mathbf{X}} -\mathbf{YP}\right\|_F^2 +\gamma_1\|\mathbf{P}\mathbf{P}^\top -\mathbf{I}\|_F^2+\gamma_2\|\mathbf{P}^\top\mathbf{P} -\mathbf{I}\|_F^2\label{eq:initialp}
\end{align}
\noindent under the constraints specified in Eqs.\eqref{eq:con1}, \eqref{eq:con2}, and \eqref{eq:acca}. We use the \emph{scipy.optimize.minimize}  solver to find the optimal $\mathbf{P}$.

\subsection{Synthetic Data Generation}
\label{sec-Datasets}
We first generate the groundtruth latent representation of the two datasets, namely $\mathbf{Z}\in\mathbb{R}^{\bar{d}\times N}$, where the columns of $\mathbf{Z}$ are  $N$  i.i.d. samples drawn from  multivariate normal distribution with zero mean and identity covariance  of size $\bar{d}\times\bar{d} $. Next, two aligned datasets $\mathbf{X}$ and $\bar{\mathbf{Y}}\in\mathbb{R}^{D_y\times N}$ are generated from their shared latent representation $\mathbf{Z}$ through two independent random projections: $\mathbf{X}=\mathbf{W} \mathbf{Z} $ and $\mathbf{Y}=\mathbf{Q}\mathbf{Z} $ where $\mathbf{W}\in\mathbb{R}^{D_x \times \bar{d}}$ and $\mathbf{Q}\in\mathbb{R}^{D_y \times \bar{d}}$. For each experiment, the groundtruth $\bar{\mathbf{P}}$ is a random permutation matrix with only one entry in each row and column to be $1$ and the rest to be $0$s. Next, we have two \emph{unaligned} datasets: $\mathbf{X}$ and $\mathbf{Y}=\bar{\mathbf{Y}}\bar{\mathbf{P}}$. The involved parameters are set as follows: $N=20$, $\bar{d}=2$, $d=7$, $D_x=15$, and $D_y=10$.

\subsection{Experimental Results}
\label{sec-Results}
\begin{figure}[!htp]
  \includegraphics[width=0.8\linewidth]{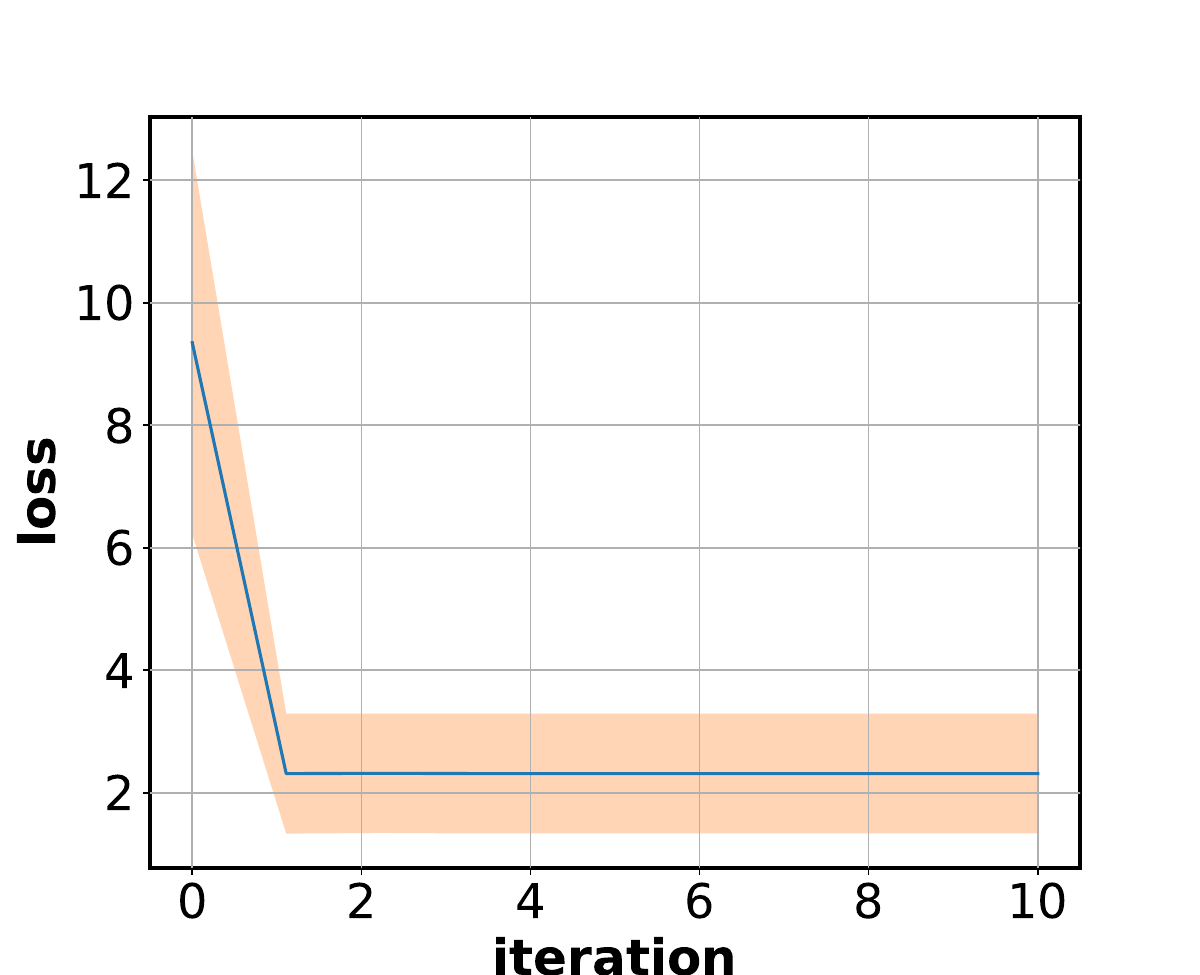}
  \caption{Loss as a function of iterations}
  \label{fig:loss}
\end{figure}

After setting the entropy upper bound hyperparameter $\lambda$ to be $0.1$, we run $10$ times of Monte Carlo experiments and report the loss of Eq. \eqref{eq:acca_loss} for each iteration in Figure \ref{fig:loss}. The curve in Figure 1 represents the average loss per iteration and the width of the shade stands for the standard derivation of the loss. Clearly, our proposed ACCA converges to a stable point using the generated synthetic data.

\begin{figure}[!htp]
  \includegraphics[width=0.8\linewidth]{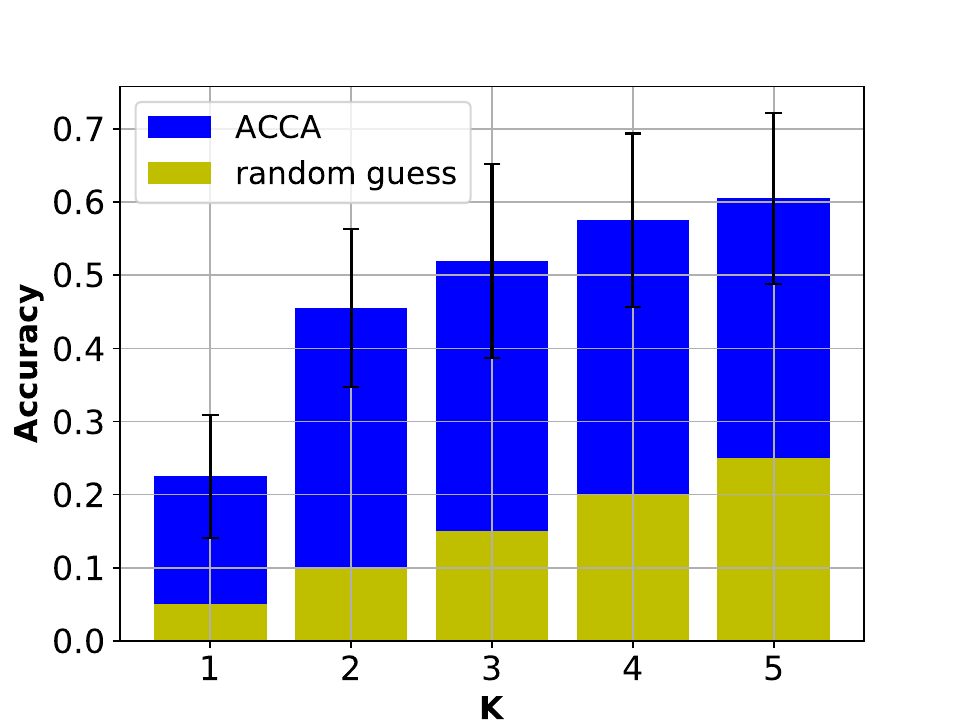}
  \caption{Top-k Accuracy of ACCA and Random guess} 
  \label{fig:topk}
\end{figure}

In Figure \ref{fig:topk}, we report the top-$k$ matching accuracy with mean and standard deviation, defined as the percentage of rows in the estimated permutation $\mathbf{P}$ whose top $k$ entries' index set includes the nonzero entry index of the true permutation  $\bar{\mathbf{P}}$, with $k=1,2,3,4,$ and $5$, in comparison with such accuracy from random guess which is $k/N$.  According to our experimental records as shown in  figure \ref{fig:topk}, it's obvious that our ACCA framework has significantly better performance in predicting the potential alignment between two datasets, than that obtained from the random guess. 

\begin{figure*}[!htp]
    \subfigure[True $\mathbf{P}$]{\includegraphics[width = 0.19\textwidth]{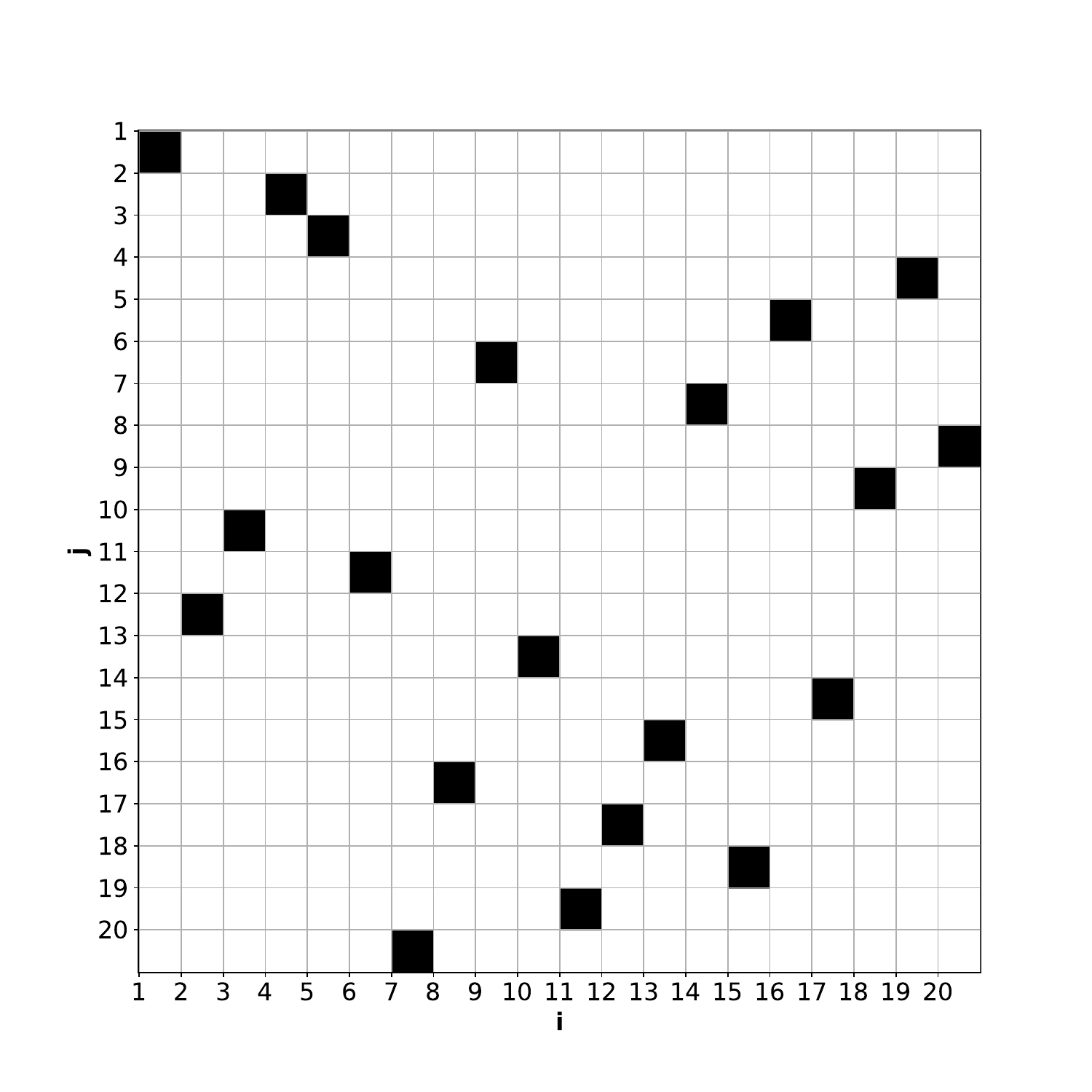}}
    \subfigure[Entropy = 0.1; top-3 acc.: 0.519]{\includegraphics[width = 0.19\textwidth]{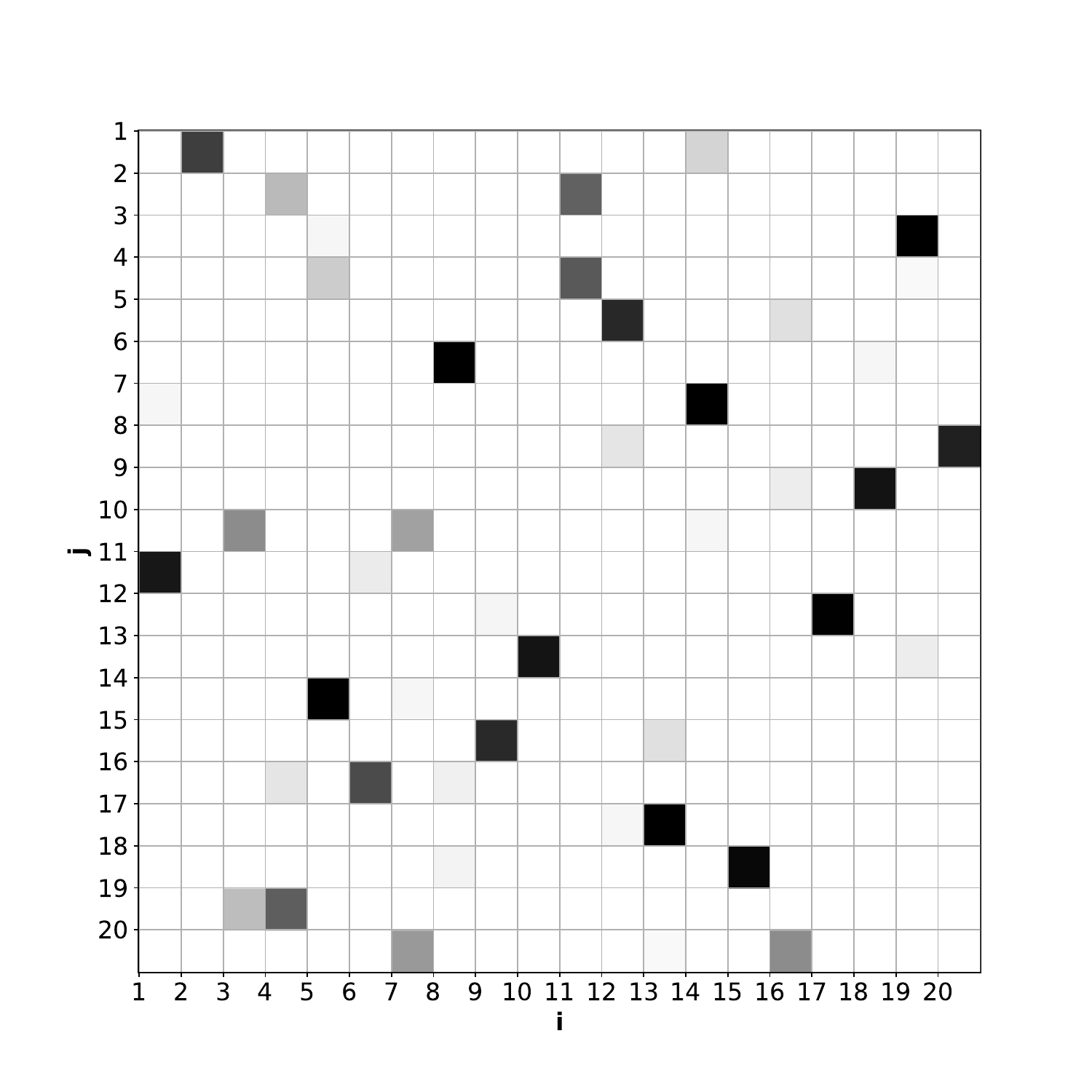}}
    \subfigure[Entropy = 0.5; top-3 acc.: 0.59]{\includegraphics[width = 0.19\textwidth]{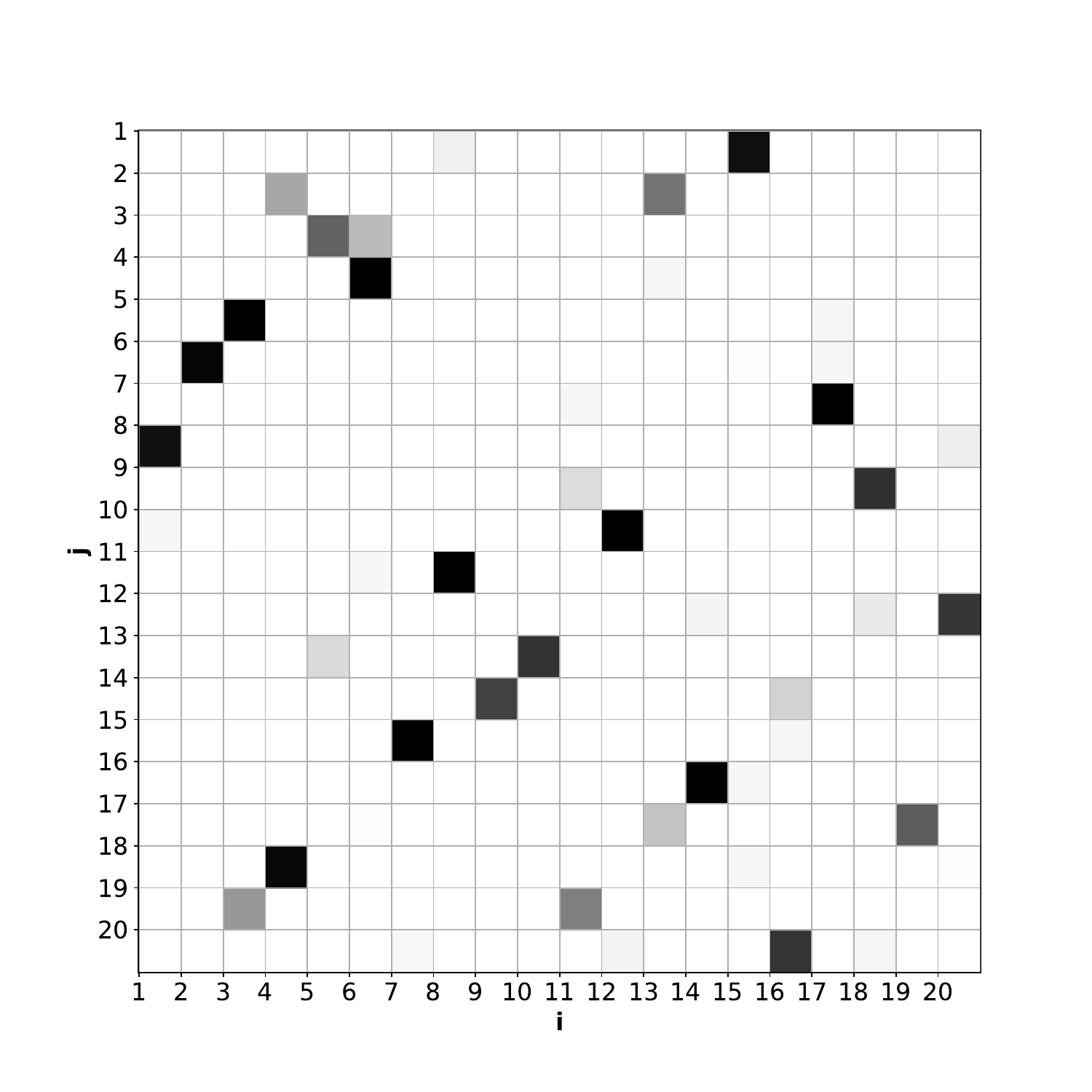}}
    \subfigure[Entropy = 1; top-3 acc.: 0.575]{\includegraphics[width = 0.19\textwidth]{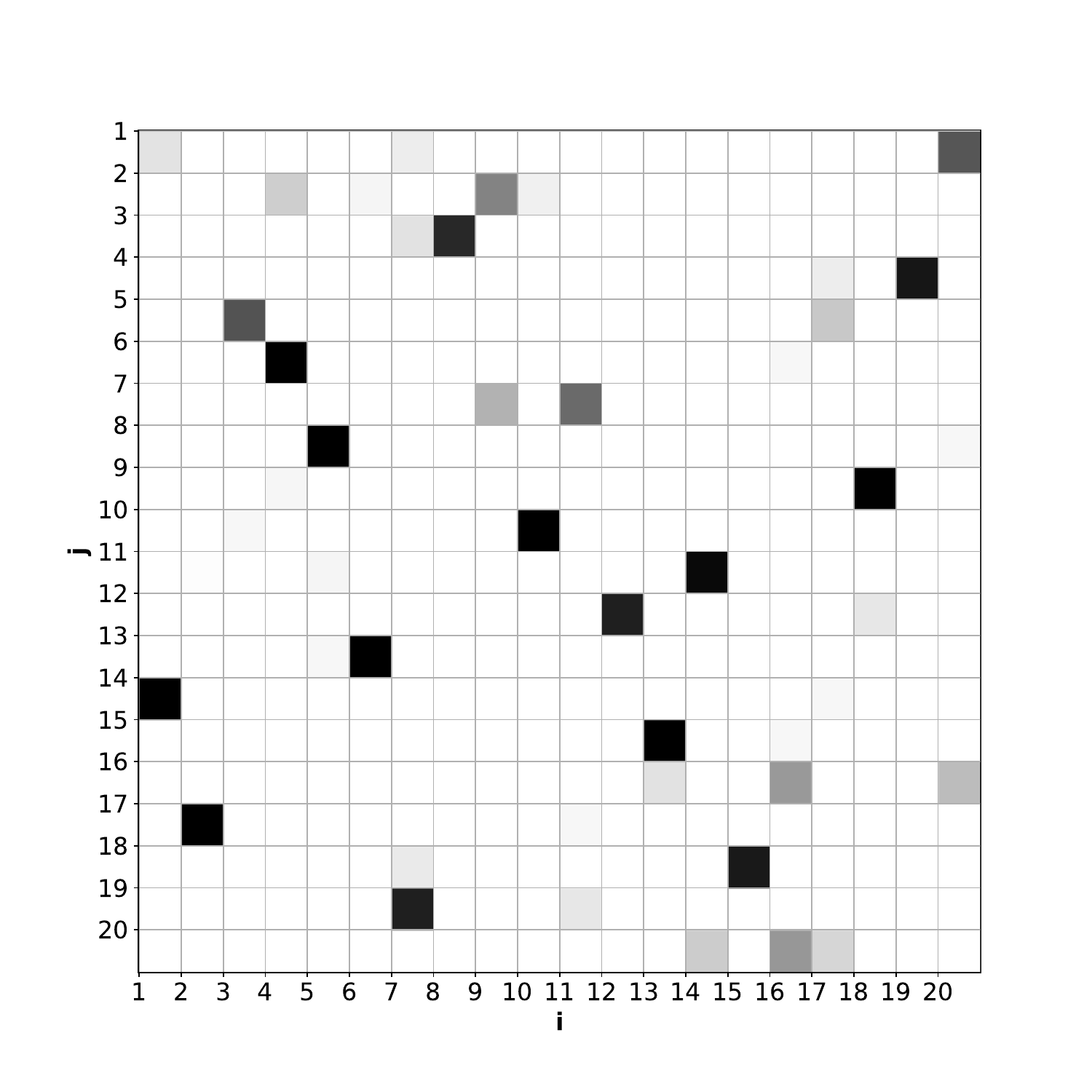}}
    \subfigure[Entropy = 2;
    top-3 acc.: 0.31]{\includegraphics[width = 0.19\textwidth]{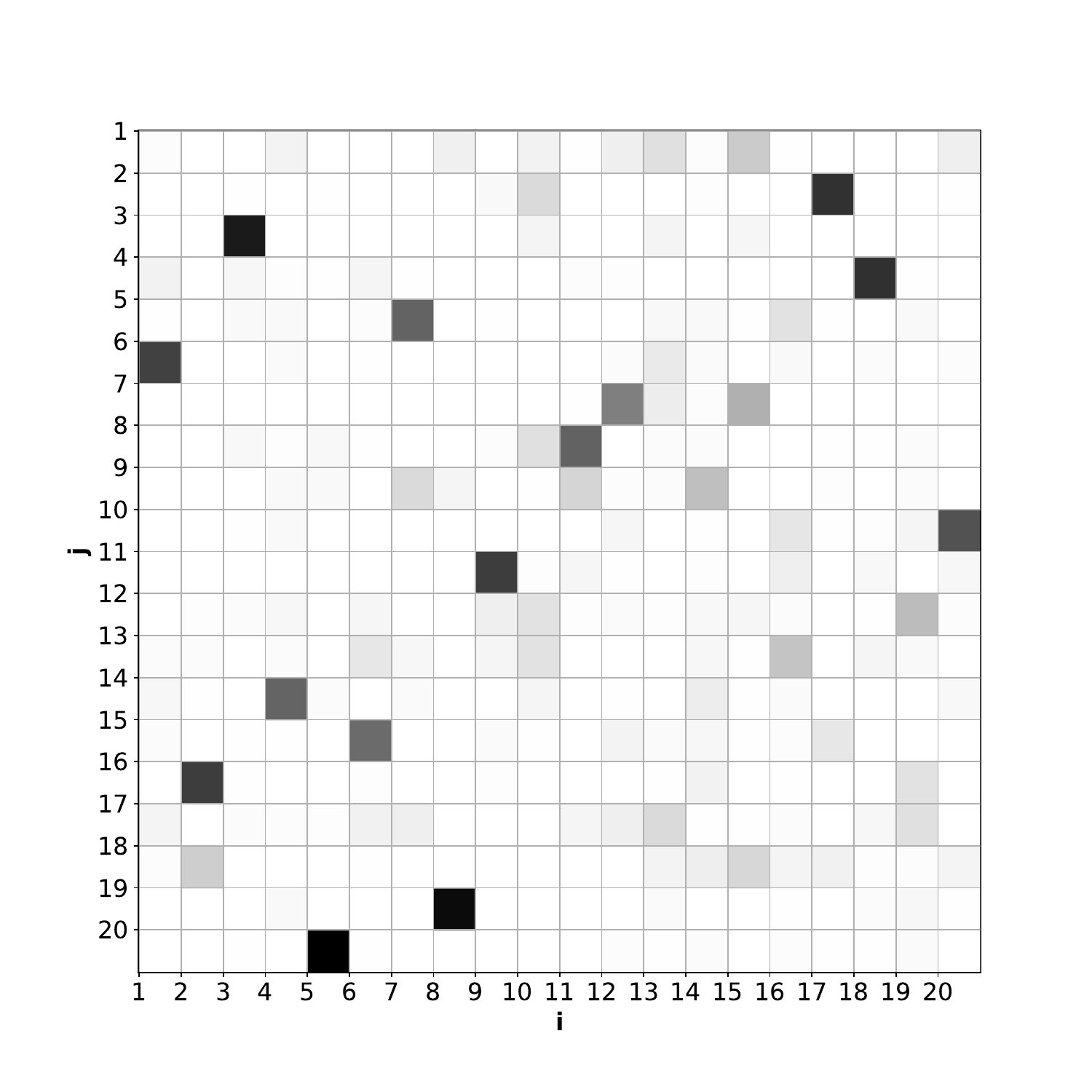}}
    \caption{Estimated alignment matrix for different Entropy bounds.}
    \label{fig:alignment_matrices}
\end{figure*}

Next, we visualize the alignment performance with respect to different values of the hyperparameter $\lambda$ in Figure \ref{fig:alignment_matrices} where we plot the real permutation matrix $\bar{\mathbf{P}}$ and the estimated $\mathbf{P}$ as gray-scale images with darker grid blocks representing higher values of the corresponding entries of $\bar{\mathbf{P}}$ or $\mathbf{P}$. As uniform distribution leads to the highest entropy, $\lambda$ can not exceed $log(N)$ (=$N \times 1/N \times log(1/N)$). With  $\lambda$ increasing, more nonzero entries are showing up in $\mathbf{P}$ as expected. With proper setup of entropy bound hyperparameter, the performance of ACCA will be further improved, with the comparison of prediction accuracies related to different entropy cases in Figure \ref{fig:alignment_matrices}. 
%($=N\times 1/N \times log(1\N)$). 

%% file: Sections/conclusion.tex
In this preliminary work we investigated the joint CCA-style embedding of multiview data and the simultaneous alignment of the embedded entities, by breaking the traditional assumption in CCA that predicates a known one-to-one matching across views. We proposed an initial formulation for Aligned Canonical Correlation Analysis (ACCA) and derived an alternating optimization algorithm that produces proof-of-concept results for the viability of this formulation. However, there is still a lot of work to be done, and we hope that our preliminary results can serve as a stepping stone to further research in this direction.

In our on-going and future work we will investigate variations of the formulation and improvements of the optimization scheme, especially as it pertains to solving for the alignment matrix, which, even though has been radically simplified compared to solving for a permutation matrix, is still a major challenge both in terms of scalability as well as in terms of finding a precise alignment matrix. Furthermore, we would like to study the alignment matrix as a graph and introduce graph-based constraints which may further improve optimization.
Finally, we will investigate connections between our proposed Aligned Canonical Correlation Analysis model and self-supervised representation learning models.